\pdfoutput=1
\documentclass[11pt]{article}
\usepackage[preprint]{acl}
\usepackage{times}
\usepackage{latexsym}
\usepackage[T1]{fontenc}
\usepackage[utf8]{inputenc}
\usepackage{microtype}
\usepackage{amsmath,amssymb,amsthm}
\usepackage{graphicx}
\usepackage{booktabs}
\usepackage{enumitem}

\newtheorem{theorem}{Theorem}
\newtheorem{proposition}{Proposition}
\newtheorem{corollary}{Corollary}
\newtheorem{lemma}{Lemma}
\theoremstyle{definition}

\theoremstyle{remark}

\newcommand{\logit}{\operatorname{logit}}
\newcommand{\PP}{\mathbb{P}}
\newcommand{\EE}{\mathbb{E}}
\newcommand{\dry}{\mathrm{DRY}}
\newcommand{\cnt}{\mathrm{cnt}}
\newcommand{\ro}{r^{\circ}}

\title{When Do Repetition Loops Lock?\\Truncation Sampling, Bounded Penalties, and a Blind Spot in DRY}

\author{Yichuan Chen \\
  School of Art and Design \\
  Zhejiang Sci-Tech University \\
  Hangzhou, China \\
  \texttt{smallboychen@gmail.com} \\\And
  Xin Wang \\
  School of Art and Design \\
  Zhejiang Sci-Tech University \\
  Hangzhou, China \\
  \texttt{wxcaup@zstu.edu.cn}}

\begin{document}
\maketitle

\begin{abstract}
Language models sometimes repeat verbatim until the generation limit.
We study the exit rather than the entrance: given that a span has already been repeated $n$ times, when is escape still possible?
Along the loop the context is deterministic, so the probability that a loop survives $m$ more repetitions is, under any sampler, a product of per-step continuation probabilities computable from one forward pass.
A lock, where this probability stays exactly one, is its limiting case.
Pure sampling never locks; top-$p$ with threshold $\pi>1/2$ locks a position exactly when the continuation probability reaches $\pi$.
A bounded penalty can leave a position locked; a penalty that outgrows the model's preference for the loop over unseen tokens guarantees escape under top-$p$.
The same analysis shows a blind spot in the widely deployed DRY sampler: its sequence breakers cap the penalty at $\lambda b^{g-1-a}$ for loops whose longest breaker-free run has $g$ tokens.
On eleven open models, 61\% of locked units with $g=6$ stay locked under DRY and 0.2\% with $g=8$.
In free generation with llama.cpp, all 199 loops that survive DRY, on six models, have $g\le 7$.
We propose a count term that grows exponentially with the number of verbatim recurrences, up to its look-back window, and closes this blind spot.
Added to DRY (DRY-C), it cuts loops in 1,800 structured generations on five models from 167 to 34, and loops reaching the length limit from 150 to 0, with reasoning accuracy and item coverage on par with DRY.
DRY's own length term costs 2.5 to 14.2 points of accuracy on GSM8K and MATH-500; used in its place, the count term keeps the default sampler's accuracy and leaves one of the 1,800 generations at the limit.
\end{abstract}

\section{Introduction}

Open-weight models can repeat a line until the token budget is exhausted, including inside reasoning, JSON and tool calls.
The phenomenon is old \citep{holtzman2020curious}, and its \emph{onset} has been studied from several angles: the self-reinforcing rise of repetition probability \citep{xu2022learning}, properties of the transition matrix \citep{fu2021theoretical}, training data \citep{li2023repetition}, and, most recently, learning errors in reasoning models \citep{pipis2025wait}.

This paper asks the complementary question.
Suppose a loop has started. Why can the model not get out, and which decoding-time interventions can in principle get it out?

Our starting observation is that, conditioned on staying in a loop, the context is a deterministic sequence.
Every sampler, including history-dependent penalties, therefore assigns a deterministic probability to ``continue the loop'' at each step, and the probability that a loop survives $m$ more repetitions is a product of such numbers (Proposition~\ref{prop:prod}).
Teacher forcing supplies the logits without sampling loop trajectories.

\textbf{Truncation can remove the exit.}
Under pure sampling every step has a positive escape probability.
Under top-$p$ with the usual threshold $\pi>1/2$, a position is \emph{locked} (the loop token is the only candidate) exactly when its raw probability $q$ satisfies $q\ge \pi$ (Theorem~\ref{thm:lock}).
Measured \emph{reinforcement curves} cross this threshold within a few repetitions. In our finite-horizon tests, 71\%--100\% of sampled units are locked through the last ten repetitions under top-$p$ 0.9, compared with none under pure sampling; the units include corpus sentences and hand-written structured spans.
Within a token budget, a per-repetition escape probability close to zero acts like a lock, and the survival product measures both.

\textbf{Penalty growth determines the available guarantee.}
For an additive logit penalty $d$ on the loop token, locking persists whenever $\logit(q)-d\ge\logit(\pi)$ (Theorem~\ref{thm:pen}).
A bounded penalty only raises the threshold at which locking sets in. Escape is guaranteed under top-$p$ once the penalty outgrows $\ln(q/\ro)$, where $\ro$ is the raw mass of tokens not yet in the context and therefore unpenalized (Corollary~\ref{cor:growing}).
A full-context frequency penalty grows fast enough as long as $\ro$ stays bounded below; presence penalties and windowed frequency penalties come with no such guarantee.

\textbf{DRY has a blind spot.}
DRY \citep{weidmann2024dry,weidmann2026dry} penalizes a token exponentially in the length of the verbatim match it would extend, and is widely used in local inference.
To protect formatting, matches may not cross \emph{sequence breakers} (newline, colon, quote, asterisk).
Breakers are densest in structured output, which is also where loops are most common: under default settings, prose prompts loop in 0\% (Qwen3-4B) and 7\% (Llama-3.1-8B) of generations and structured prompts in 21\% and 16\%, and in 1\% to 34\% on three further models.
We show that for a loop whose longest breaker-free run has $g$ tokens, the penalty on the loop token is at most $\lambda b^{g-1-a}$ (Proposition~\ref{prop:dry}): on such loops DRY is a bounded penalty.
In a dose--response experiment, locks persist under DRY up to $g=6$ and almost vanish from $g=8$; the unmodified llama.cpp sampler keeps breaker-dense units locked; and in free generation every one of the 199 loops that appear with DRY enabled has $g\le7$.

\textbf{A count term closes it.}
We propose a \emph{count term} that penalizes how many times the same continuation has already occurred, and prove that it grows exponentially along every loop, breakers or not, up to its look-back window (Proposition~\ref{prop:count}).
Added to DRY (DRY-C), it eliminates run-to-the-limit loops on five models from four families, with no loss of reasoning accuracy relative to DRY and item coverage within 2.3 points.
DRY itself, however, lowers reasoning accuracy by up to 14.2 points, through its length term.
Used in place of that term, the count term keeps the default sampler's reasoning accuracy and leaves 1 of 1,800 structured generations at the limit, at the cost of fewer new items in itemized output.
The blind spot is closed either way; what remains is a choice about DRY's length term.

\section{Related Work}

\paragraph{Why loops start.}
\citet{holtzman2020curious} documented degenerate repetition under likelihood-maximizing decoding and introduced nucleus sampling.
\citet{fu2021theoretical} analyzed repetition through a Markov model of the transition matrix.
\citet{xu2022learning} measured a \emph{self-reinforcement effect}: the probability of repeating a sentence rises with the number of previous repetitions; our reinforcement curves extend this measurement to current instruction-tuned and reasoning models and to 200 repetitions.
\citet{li2023repetition} traced repetition to training data, \citet{yao2025repeat} and \citet{mahaut2025repetitions} to internal features and mechanisms.
\citet{pipis2025wait} explain why reasoning models enter loops through learning errors and study the role of temperature, keeping each model's default truncation and explicitly setting penalties aside.
\citet{lee2026focus} separate loop entry from loop persistence in pruned models, relate persistence to the escape mass that the sampling set leaves to alternatives, and address both with fine-tuning objectives.
We take the loop as given, work at decoding time, and characterize when that escape mass is exactly zero; these views are complementary.

\paragraph{Truncation samplers.}
Top-$p$ \citep{holtzman2020curious}, min-$p$ \citep{nguyen2024minp}, typical sampling \citep{meister2023locally}, and related truncation rules \citep{hewitt2022truncation,finlayson2024closing,basu2021mirostat} are motivated by the unreliability of the distribution's tail.
We point out a side effect: truncation turns a near-certain continuation into a certain one.

\paragraph{Repetition penalties.}
The multiplicative penalty of \citet{keskar2019ctrl}, presence and frequency penalties, training-time remedies \citep{welleck2020neural,xu2022learning}, contrastive search \citep{su2022contrastive}, and penalty decoding with a forgetting window \citep{zhu2023penalty} all discourage reuse of tokens; hard $n$-gram blocking \citep{paulus2018deep,klein2017opennmt} forbids it outright.
\citet{ginart2025lz} derive a penalty from LZ77 code lengths.
\citet{wang2025solving} compare remedies in a production setting.
Grammar-constrained decoding \citep{willard2023efficient,dong2024xgrammar} guarantees well-formed structured output but does not stop a valid record from being repeated.
DRY \citep{weidmann2024dry} acts on repeated \emph{continuations} rather than repeated tokens; \citet{weidmann2026dry} evaluate it at scale, report roughly halved loop rates, and show that removing sequence breakers damages formatting while lowering their loop metric slightly further, treating breakers as a structure-preserving benefit.
We characterize the cap, the loops it can leave locked, and a count-based modification that retains breakers. Further decoding and monitoring approaches are discussed in Appendix~\ref{app:additional}.

\section{Setup}\label{sec:setup}

Let $V$ be the vocabulary.
Fix a prefix and a \emph{loop unit} $u=(u_1,\dots,u_L)\in V^L$.
In \emph{state} $(n,l)$, with $n\ge0$ and $1\le l\le L$, the context is the prefix, followed by $n$ complete copies of $u$, followed by $u_1,\dots,u_{l-1}$; the model is about to produce the $l$-th token of copy $n+1$.
Write $p_{n,l}$ for the model's next-token distribution in this state, and
\[
q_{n,l}=p_{n,l}(u_l),\qquad \tilde q_{n,l}=\max_{v\ne u_l}p_{n,l}(v)
\]
for the raw probability of continuing the loop and of the strongest rival.
We drop subscripts when no confusion arises. Throughout, $V$ is finite, logits are finite, temperatures satisfy $0<\tau<\infty$ (greedy decoding is treated separately), $\lambda>0$, and the lengths and counts $a$, $a'$, $c_0$ are positive integers.

A \emph{sampler} $S$ maps the history, meaning the tokens so far together with the model's logits at every step (which the tokens determine), to a distribution over the next token, possibly using fresh randomness at each step.
Along the loop the history is determined by $(n,l)$, so the probability that $S$ emits $u_l$ is a deterministic number $q^S_{n,l}$; this includes samplers whose internal state is a function of the history, such as Mirostat.
Let $Q_n=\prod_{l=1}^{L}q^S_{n,l}$ be the probability of completing one more copy.
State $(n,l)$ is \emph{locked} if $q^S_{n,l}=1$; repetition $n$ is locked if $Q_n=1$.
Starting with $k$ copies in context, let $T_k\in\{0,1,\dots\}\cup\{\infty\}$ be the number of further complete copies produced before the first deviation.

\begin{proposition}[Survival]\label{prop:prod}
For any sampler, $\PP(T_k\ge m)=\prod_{j=k}^{k+m-1}Q_j$ and $\PP(T_k=\infty)=\prod_{j\ge k}Q_j$.
\end{proposition}

The proof is the chain rule (Appendix~\ref{app:proofs}).
Its practical content is that loop survival under any sampler is computable exactly from the logits along the repetition path.
We distinguish three notions: a position is \emph{not locked} ($q^S<1$); there is \emph{no permanent lock} (no $n_0$ beyond which every repetition is locked); and the loop \emph{ends almost surely} ($\PP(T_k=\infty)=0$).

\section{Truncation Can Lock Loop Positions}\label{sec:lock}

\begin{proposition}[Pure sampling]\label{prop:pure}
If $S$ only rescales temperature, no state is locked.
Moreover $\PP(T_k=\infty)>0$ if and only if $\sum_{j\ge k}(1-Q_j)<\infty$, and if $Q_j\le\bar Q<1$ for all $j\ge k$ then $\EE[T_k]\le\bar Q/(1-\bar Q)$.
\end{proposition}

Truncation changes the picture qualitatively.
We use the standard definitions.
Top-$p$ with threshold $\pi\in(\tfrac12,1)$ keeps token $v$ iff the total probability of tokens strictly more probable than $v$ is below $\pi$.
Min-$p$ with ratio $\beta\in(0,1)$ keeps $v$ iff $p(v)\ge\beta\max_w p(w)$.
Greedy decoding keeps the most probable token.
We assume no exact ties among the largest probabilities; implementations break ties by token order.

\begin{theorem}[Lock criteria]\label{thm:lock}
For top-$p$ let $\pi\in(1/2,1)$; for min-$p$ let $\beta\in(0,1)$. State $(n,l)$ is locked
\begin{enumerate} [label=(\alph*),nosep]
\item under top-$p$ iff $q\ge\pi$;
\item under min-$p$ iff $\ln q-\ln\tilde q>\ln(1/\beta)$;
\item under greedy decoding iff $q>\tilde q$.
\end{enumerate}
If the condition holds at every position for all $n\ge n_0$, then $\PP(T_{n_0}=\infty)=1$.
\end{theorem}

Whether a position locks can therefore be read off the \emph{reinforcement curve} $n\mapsto q_{n,l}$.
Top-$k$ applied before top-$p$ (the default order in llama.cpp) renormalizes over the $k$ retained tokens: the state is then locked iff $u_l$ is retained among the top $k$ tokens and $q/m_k\ge\pi$, where $m_k$ is their total raw probability. For a retained loop token, this renormalization can only make locking easier.

\paragraph{Temperature and sampler order.}
Temperature $\tau$ maps $p$ to $p^{(\tau)}\propto p^{1/\tau}$.
Its effect depends on where it sits in the chain.

\begin{proposition}[Temperature]\label{prop:temp}
(A) If truncation precedes temperature (the default order in llama.cpp), the criteria of Theorem~\ref{thm:lock} are unchanged.
(B) If temperature precedes truncation (the default in \texttt{transformers}) and $\tau\le1$, then for top-$p$
\begin{align*}
\logit(q)\ge\tau\logit(\pi)\ &\Rightarrow\ \text{locked}\\
&\Rightarrow\ \ln\tfrac{q}{\tilde q}\ge\tau\logit(\pi),
\end{align*}
and for min-$p$, for every $\tau>0$, the state is locked iff $\ln q-\ln\tilde q>\tau\ln(1/\beta)$.
\end{proposition}

With $\pi=0.9$ and $\tau=0.7$ in order (B), the sufficient threshold drops from $q\ge0.9$ to $q\ge0.823$: lowering the temperature also lowers the bar for becoming trapped.
The case $\tau>1$ is treated in Appendix~\ref{app:proofs}.

\section{Penalties: Bounded versus Growing}\label{sec:pen}

An \emph{additive penalty} subtracts $d(v)\ge0$ from the logit of $v$ before truncation, where $d$ may depend on the history.
All penalties considered here leave unseen tokens untouched.
Let $d=d(u_l)$ and let $\ro$ be the raw probability mass of tokens absent from the context; we assume this untouched set is nonempty, so $\ro>0$.

\begin{theorem}[Penalized continuation]\label{thm:pen}
Assume $n\ge1$, so the loop token has appeared in the context, and $\ro>0$. After the penalty, the loop token has probability
\[
q'=\frac{q\,e^{-d}}{q\,e^{-d}+\sum_{v\ne u_l}p(v)e^{-d(v)}},
\]
with $\frac{qe^{-d}}{qe^{-d}+1-q}\le q'\le\frac{qe^{-d}}{qe^{-d}+\ro}$.
Hence, with top-$p$ after the penalty,
\begingroup\small
\begin{align*}
\logit(q)-d\ge\logit(\pi)\ &\Rightarrow\ \text{locked}\\
&\Rightarrow\ \ln\tfrac{q}{\ro}-d\ge\logit(\pi).
\end{align*}
\endgroup
With min-$p$ after the penalty, the state is locked iff
\begin{multline*}
\ln q-d-\max_{v\ne u_l}(\ln p(v)-d(v))
>\ln(1/\beta).
\end{multline*}
\end{theorem}

\begin{corollary}[Bounded penalties]\label{cor:bounded}
With top-$p$ ($\pi\in(1/2,1)$) after the penalty, if $d_{n,l}\le D$ for all states, then every state with  $q_{n,l}\ge\sigma(\logit(\pi)+D)$ is locked, where $\sigma$ is the logistic function.
\end{corollary}

A bounded penalty thus moves the threshold from $\pi$ to $\sigma(\logit(\pi)+D)$; for $\pi=0.9$ and $D=1,2,3$ this is $0.961$, $0.985$, $0.9945$.
It cannot remove locking from a model whose reinforcement curve saturates above that level.

\begin{corollary}[Growing penalties]\label{cor:growing}
With top-$p$ ($\pi\in(1/2,1)$) after the penalty, if $d_{n,l}-\ln(q_{n,l}/\ro_{n,l})\to\infty$ at some position $l$, then  $Q_n\to0$ and the loop ends almost surely.
This holds in particular if $d_{n,l}\to\infty$ and $\ro_{n,l}\ge\rho>0$.
\end{corollary}

\begin{table}[t]
\centering\small\setlength{\tabcolsep}{3pt}
\begin{tabular}{@{}lll@{}}
\toprule
Penalty & $d$ on loop token & Class \\
\midrule
Presence $\beta_p$ & $\beta_p$ & bounded \\
Frequency $\alpha$, full context & $\ge\alpha n$ & linear \\
Frequency $\alpha$, window $W$ & $\le\alpha W$ & bounded \\
Multiplicative $\varrho$ & $z(1-\tfrac1\varrho)^{\ddagger}$ & bounded$^\dagger$ \\
DRY, no breaker in unit & $\ge\lambda b^{(n-1)L-a}$ & exponential \\
DRY, breaker-free run $g$ & $\le\lambda b^{g-1-a}$ & bounded \\
DRY-C & $\ge\lambda b_c^{\,n-\lceil a'/L\rceil-c_0}$ & exponential \\
\bottomrule
\end{tabular}
\caption{Penalties along a loop. Exponential rows assume $W=\infty$; finite windows cap their growth. $z$: loop-token logit. $^\ddagger$For $z>0$; for $z<0$ it is $|z|(\varrho-1)$. The DRY lower bound needs $(n-1)L\ge a$; the DRY-C lower bound needs $n-\lceil a'/L\rceil\ge c_0$. $^\dagger$Bounded when $z$ is; multiplicative penalties depend on the logit origin.}
\label{tab:class}
\end{table}

Table~\ref{tab:class} classifies common penalties.
Two entries deserve comment.
First, the \emph{look-back window} decides the class: over the full context the frequency penalty grows linearly with $n$, but restricted to the last $W$ tokens (64 by default in llama.cpp) a token appears at most $W$ times; once the window lies entirely within the looping region, the sharper bound is $m\lceil W/L\rceil$, where $m$ is its count per unit.
Every implementation counts over a finite look-back, at most the context length, which bounds every additive penalty; what matters is whether the cap exceeds the lock margin for the loop lengths that occur.
The frequency cap grows linearly in $W/L$; the DRY-C cap grows exponentially (Section~\ref{sec:dry}).
For a linear penalty $d\ge\alpha n$ and a finite supremum of the margin, Theorem~\ref{thm:pen} gives an explicit unlock time under top-$p$: no state is locked beyond  $n^{*}=\sup_{n,l}\bigl(\ln(q/\ro)-\logit(\pi)\bigr)/\alpha$.
Second, the multiplicative penalty divides positive logits and multiplies negative ones, so its strength depends on where the model places the zero of its logits, a quantity with no meaning for the distribution; \citet{hollows2026gauge} measure this effect and its damage to structured output.

\section{DRY and Its Sequence Breakers}\label{sec:dry}

\subsection{The penalty cap}

We state DRY as implemented in llama.cpp \citep{gerganov2023llamacpp}.
Parameters are a multiplier $\lambda$, a base $b>1$, and an allowed length $a$ (defaults $0.8$, $1.75$, $2$).
Let $B\subset V$ be the set of \emph{breaker tokens}, those whose text contains a newline, colon, quotation mark, or asterisk (with these default single-character breakers every breaker occurrence is a single token).
For context $x_1,\dots,x_t$, choose a look-back window $W$ and put $h=\max\{1,t-W+1\}$ ($h=1$ if $W=\infty$). Let $\kappa$ be the trailing non-breaker length and $M_j$ the common-suffix length of $x_{1:j}$ and $x_{1:t}$.
The windowed match length is
\[
M(v)=\max_{\substack{h\le j<t\\x_{j+1}=v}}\min\{M_j,\kappa,j-h+1\},
\]
with an empty maximum equal to zero, and
\[
d_{\dry}(v)=\begin{cases}\lambda\,b^{M(v)-a} & M(v)\ge a,\ v\notin B,\\ 0 & \text{otherwise.}\end{cases}
\]
Breakers enter twice: they truncate the match, and breaker tokens themselves are never penalized.

For a unit containing at least one breaker token, arrange $u$ on a circle and let $g$ be the largest number of consecutive non-breaker tokens.

\begin{proposition}[Penalty cap]\label{prop:dry}
For every state $(n,l)$ with $n\ge1$,
\[
d_{\dry}(u_l)\le D_g:=\begin{cases}\lambda\,b^{g-1-a} & g\ge a+1,\\ 0 & g\le a,\end{cases}
\]
and the bound is attained for $g\ge a+1$, $n\ge2$, and $W\ge L+g-1$.
\end{proposition}

\begin{proof}[Proof sketch]
If $u_l\in B$ the penalty is zero.
Otherwise the $\kappa$ trailing non-breaker tokens together with $u_l$ form $\kappa+1$ consecutive non-breaker tokens on the circle, so $\kappa\le g-1$ and $M(u_l)\le g-1$.
\end{proof}

\begin{corollary}[Blind spot]\label{cor:blind}
With top-$p$ after DRY, every state of such a loop with $n\ge1$ and $q_{n,l}\ge\sigma(\logit(\pi)+D_g)$ is locked.
\end{corollary}

With default parameters, $D_g=0.8,\ 1.4,\ 2.45,\ 4.29,\ 7.50,\ 13.1$ for $g=3,\dots,8$, so under top-$p$ 0.9 a lock is guaranteed once the log-odds reach $3.0$, $3.6$, $4.7$, $6.5$, $9.7$, $15.3$.
After 200 repetitions the median log-odds exceed the $g=6$ threshold on ten of our eleven models, the $g=7$ threshold on five and the $g=8$ threshold on two (Section~\ref{sec:curves}); a lock needs every position of the unit, so the guarantee bites on loops with short breaker-free runs---JSON records, key--value lines, quoted dialogue, short lines.
For larger $g$ the cap exceeds typical margins, and whether a given unit unlocks is an empirical question that Section~\ref{sec:dose} answers.
Without breakers and with unlimited look-back, the match length is at least $(n-1)L$; a finite window eventually caps this growth.

\subsection{DRY-C: a count term}

Match length is capped by breakers, but a loop has a second signature that breakers do not affect: the same continuation recurs \emph{many times}.
Let $a'\ge1$ (minimum length), $c_0\ge1$ (allowed recurrences) and $b_c>1$.
Let $J_W=\{j:h+a'-1\le j<t\}$. Define
\[
c(v)=\#\{j\in J_W:\ x_{j+1}=v,\ M_j\ge a'\},
\]
that is, how many earlier positions in the window are preceded by the same $a'$ tokens as the current one, verbatim and ignoring breakers, and are followed by $v$ ($W=4096$ here).
The count term is $d_{\cnt}(v)=\lambda\,b_c^{\,c(v)-c_0}$ if $c(v)\ge c_0$ and $0$ otherwise; it applies to breaker tokens as well, and the total penalty is $\max\{d_{\dry},d_{\cnt}\}$. We call the combined sampler DRY-C (C for count), and the count term used without DRY's length term \emph{count only}.

\begin{proposition}[Growth along every loop]\label{prop:count}
For any unit $u$, write $N_W=\lfloor (W-a')/L\rfloor$. In every state with $\min(n-\lceil a'/L\rceil,N_W)\ge c_0$,
\[
d_{\cnt}(u_l)\ \ge\ \lambda\,b_c^{\,\min(n-\lceil a'/L\rceil,\,N_W)-c_0}.
\]
With an unlimited window ($N_W=\infty$) the bound grows without limit in $n$.
\end{proposition}

\begin{proof}[Proof sketch]
The same position in each of the $n$ earlier copies is followed by $u_l$; for the $k$-th most recent one, at least $(n-k)L$ preceding tokens agree with the current suffix.
Those with $(n-k)L\ge a'$ whose matching span lies in the window are counted.
\end{proof}

\begin{corollary}\label{cor:count}
Let DRY-C be followed by top-$p$, or by top-$k$ ($k\ge2$) and then top-$p$ as in llama.cpp. Write $\Delta=\sup_{n,l}\bigl(\ln(q/\ro)-\logit(\pi)\bigr)$ in the first case and $\Delta=\sup_{n,l}\bigl(\ln(q/r^{*})-\logit(\pi)\bigr)$ in the second, where $r^{*}$ is the largest raw probability of a token absent from the context; assume $\Delta<\infty$ and put $\delta=\log_{b_c}(\max\{\Delta,\lambda\}/\lambda)\ge0$.
With an unlimited window, no state is locked for $n>\lceil a'/L\rceil+c_0+\delta$.
With a window of $W$ tokens the same holds whenever $N_W>c_0+\delta$.
\end{corollary}

The unlock time is logarithmic in $\Delta$, compared with linear for the frequency penalty. Both results use only $d_{\cnt}$, so they hold for the count term alone.
The largest margin $\ln(q/\ro)-\logit(0.9)$ we measured is $\Delta=37.6$ (Qwen3-8B); with $W=4096$ and our parameters, the window condition then holds for loop units of up to 678 tokens, and our detector looks for periods of up to 400.
We use $a'=24$, $c_0=2$, and $b_c=3$ unless stated otherwise.

\section{Experiments}\label{sec:exp}

\subsection{Setup}

\paragraph{Models.}
Eleven instruction-tuned models from 1.5B to 9B parameters: Qwen2.5 1.5B and 7B \citep{qwen25}; Qwen3 1.7B, 4B and 8B \citep{qwen3}; Llama 3.2 3B and 3.1 8B \citep{llama3}; Gemma 2 2B and 9B \citep{gemma2}; Phi-4-mini \citep{phi4mini}; and the DeepSeek-R1 distillation of Qwen 7B \citep{deepseekr1}.

\paragraph{Teacher-forced measurement.}
We sample 150 English sentences from the WikiText-103 test set \citep{merity2017pointer} and 150 Chinese sentences from Chinese Wikipedia, stratified by length, and add 22 hand-written units (dialogue lines, list items, code).
Each unit is repeated up to 200 times after the beginning-of-sequence token, in an inline variant (joined by a space) and a line variant (terminated by a newline), keeping units whose tokenization is stable under repetition.
One forward pass with key--value caching yields $q_{n,l}$, $\tilde q_{n,l}$ and $\ro_{n,l}$ at every position; softmax is evaluated in 64-bit precision.
Given the logits, the continuation probability under any sampler configuration is computed exactly, with penalties applied before truncation as in llama.cpp.
Our DRY computation follows the definition of Section~\ref{sec:dry} and was validated position by position against the llama.cpp sampler (Appendix~\ref{app:sim}).

\paragraph{Free generation.}
We run llama.cpp with its own samplers on GGUF quantizations of Qwen3-4B, Llama-3.1-8B and Qwen3.6-35B-A3B, and on Qwen3-8B, Gemma-2-9B and Phi-4-mini for the structured prompts.
Prompt set~1 has 20 prompts (reasoning, enumeration and eight structured-output tasks); prompt set~2 has 24 structured-output prompts (JSON, tables, YAML, CSV, logs, lists, dialogue) and 12 prose-writing prompts.
``Default'' denotes the sampler recommended by each model's developers (Qwen: $\tau=0.6$, top-$p$ 0.95, top-$k$ 20; Llama: $\tau=0.6$, top-$p$ 0.9; Gemma-2-9B and Phi-4-mini: $\tau=0.6$, top-$p$ 0.95), with at most 3000 new tokens.
A generation contains a \emph{loop} if some span of at least 4 tokens occurs at least 5 times consecutively, covering at least 60 tokens.
Details are in Appendix~\ref{app:details}.

\subsection{Reinforcement curves}\label{sec:curves}

\begin{figure*}[t]
\centering
\includegraphics[width=0.92\textwidth]{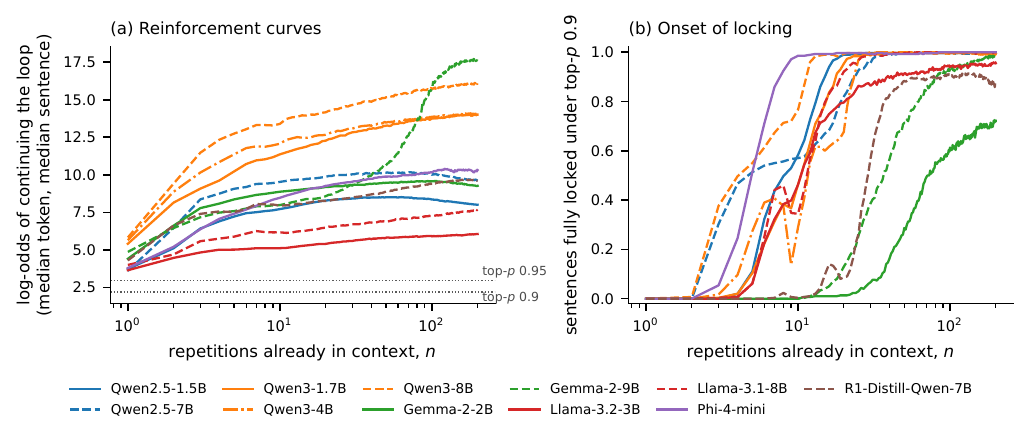}
\caption{(a) Log-odds after $n$ repetitions: median over positions within each inline corpus sentence, then median over sentences. Dotted lines mark the lock thresholds of top-$p$ 0.9 and 0.95. (b) Fraction of sentences in which every position satisfies $q\ge0.9$ at repetition $n$. At large $n$, only sentences whose context fits enter (at least 31 per point).}
\label{fig:curves}
\end{figure*}

In Figure~\ref{fig:curves}(a) the log-odds (median over positions, then over sentences) climb steeply over the first few repetitions.
Most then keep rising slowly (three dip by at most 0.6 after 60 to 110 repetitions), reaching 6 (Llama-3.2-3B) to 18 (Gemma-2-9B) after 200 repetitions; the copying behind this rise is usually attributed to induction heads \citep{olsson2022context}.
Every curve sits above the top-$p$ thresholds from the first repetition onward for the median token; locking requires this at \emph{every} position, which happens for half of the sentences after 5 to 71 repetitions (Figure~\ref{fig:curves}(b)).

\subsection{Which settings lock}

\begin{table*}[t]
\centering\small
\setlength{\tabcolsep}{4pt}
\begin{tabular}{@{}lcccccccccccc@{}}
\toprule
 & & \multicolumn{3}{c}{Truncation only} & \multicolumn{4}{c}{top-$p$ 0.9 + bounded penalty} & \multicolumn{4}{c}{top-$p$ 0.9 + other penalties} \\
\cmidrule(lr){3-5}\cmidrule(lr){6-9}\cmidrule(l){10-13}
Model & pure & top-$p$ & min-$p$ & $\tau$=0.7 & pres. & freq.$_{64}$ & rep.$_{64}$ & rep.$_{64}$ & freq. & DRY & DRY-C & count \\
 & & 0.9 & 0.1 & +top-$p$ & 1.0 & 0.5 & 1.1 & 1.3 & 0.1 & & & only \\
\midrule
Qwen2.5-1.5B & 0 & 1.00 & 1.00 & 1.00 & 1.00 & .96 & .80 & 0 & 0 & .02 & 0 & 0 \\
Qwen2.5-7B & 0 & .99 & .99 & .99 & .98 & .95 & .83 & 0 & 0 & .02 & 0 & 0 \\
Qwen3-1.7B & 0 & .99 & .99 & .99 & .99 & .99 & .98 & .21 & 0 & .02 & 0 & 0 \\
Qwen3-4B & 0 & 1.00 & 1.00 & 1.00 & .99 & .99 & .98 & .01 & 0 & .02 & 0 & 0 \\
Qwen3-8B & 0 & .99 & .99 & 1.00 & .99 & .99 & .99 & .35 & 0 & .02 & 0 & 0 \\
R1-Distill-7B & 0 & .88 & .91 & .95 & .79 & .70 & .50 & 0 & 0 & .01 & 0 & 0 \\
Llama-3.2-3B & 0 & .95 & .97 & 1.00 & .77 & .56 & .47 & 0 & 0 & .01 & 0 & 0 \\
Llama-3.1-8B & 0 & 1.00 & 1.00 & 1.00 & 1.00 & .90 & .89 & 0 & 0 & .01 & 0 & 0 \\
Gemma-2-2B & 0 & .71 & .85 & .89 & .19 & .15 & .07 & 0 & 0 & .01 & 0 & 0 \\
Gemma-2-9B & 0 & .95 & .96 & .96 & .77 & .67 & .31 & .01 & 0 & .01 & 0 & 0 \\
Phi-4-mini & 0 & .99 & .99 & 1.00 & .95 & .95 & .53 & 0 & 0 & 0 & 0 & 0 \\
\bottomrule
\end{tabular}
\caption{Fraction of units locked through the last ten measured repetitions (150 sampled units: corpus and hand-written spans). Top-$p$, DRY and DRY-C come from a second run in which DRY and DRY-C use a 4,096-token look-back (Appendix~\ref{app:additional}), and count only from a third run with the same look-back; the other columns come from the first run. Subscript 64: penalty window. Temperature precedes top-$p$ in the $\tau$=0.7 column. Phi-4 is limited to 4,000 tokens throughout.}
\label{tab:lock}
\end{table*}

\paragraph{Exactness of the criteria.}
On four models (Qwen2.5-1.5B and 7B, Qwen3-4B, Llama-3.2-3B; about 160{,}000 positions each) the criteria of Theorem~\ref{thm:lock} agree with the exactly computed lock status at every position for top-$p$ 0.9 and 0.95 and min-$p$ 0.1 and 0.05.
The sufficient and necessary conditions of Proposition~\ref{prop:temp}(B) and Theorem~\ref{thm:pen} have no violations, and the sufficient conditions cover over 99\% of locked positions (Appendix~\ref{app:verify}).

\paragraph{Samplers.}
Table~\ref{tab:lock} reports which units stay locked through the last ten measured repetitions.
Pure sampling locks nothing.
Top-$p$ 0.9 locks 71\%--100\% of units through that interval.
Among penalties, the split described in Section~\ref{sec:pen} holds: in ten of the eleven models, presence 1.0 (77\%--100\%) and windowed frequency 0.5 (56\%--99\%) leave most units locked, and the mild multiplicative penalty leaves 31\%--99\%; a strong one removes most locks (0\%--35\% remain); and the full-context frequency penalty leaves none locked through the final ten repetitions.
Gemma-2-2B is the exception: the same penalties unlock most of its units (19\%, 15\% and 7\% stay locked).
Figure~\ref{fig:pen} puts the penalty classes side by side against the measured lock margin.

\paragraph{Locked versus long.}
Loops also reach the limit without a lock.
At $\tau=0.6$ with no truncation nothing locks, yet 53 (Qwen3-4B) and 40 (Llama-3.1-8B) of 360 structured generations loop to the limit, against 59 and 54 with the default sampler (24 and 0 at $\tau=1$). Within a 3,000-token budget, a small positive escape probability acts like a lock (Proposition~\ref{prop:pure}).

\subsection{The DRY blind spot}\label{sec:dose}

\paragraph{Dose--response.}
To vary the breaker-free run length while holding content fixed, we take the first 24 tokens of 40 sentences per model and insert a colon (a breaker) after every $g$ tokens, for nominal $g\in\{2,3,4,6,8,12,24\}$; as a control we insert a semicolon (not a breaker). Retokenization can change the actual gap; Appendix~\ref{app:additional} reports both groupings.

\begin{figure}[t]
\centering
\includegraphics[width=0.95\columnwidth]{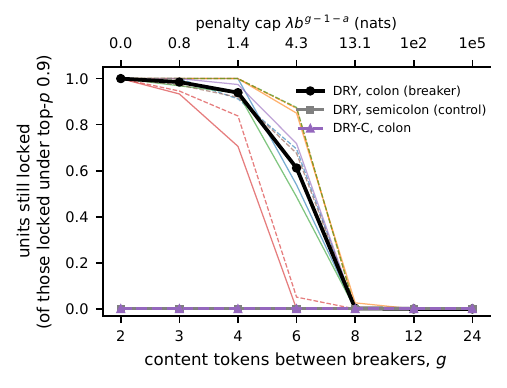}
\caption{Fraction of units that remain locked under DRY + top-$p$ 0.9, among units locked under top-$p$ alone. Thin lines: individual models; thick line: pooled over eleven models. Top axis: $D_g$ evaluated at the nominal gap; actual-gap regrouping is in Appendix~\ref{app:additional}.}
\label{fig:dose}
\end{figure}

Figure~\ref{fig:dose} shows a sharp transition.
Pooled over models, DRY leaves 387/387 units locked at $g=2$, 392/398 at $g=3$, 385/410 at $g=4$, 249/407 (61\%) at $g=6$, and 1/415 (0.2\%) at $g=8$, and none beyond.
With semicolons, none of the 2,843 units locked under top-$p$ alone remain locked under DRY.
The models that unlock earliest at $g=6$ are the two Llama models, whose reinforcement curves are the lowest in Figure~\ref{fig:curves}, in line with Corollary~\ref{cor:blind}.
Among the units of Table~\ref{tab:lock}, the 21 that stay locked under DRY all have $5\le g\le7$ (15 hand-written dialogue lines and 6 English sentences), and DRY-C leaves none.

\paragraph{The real sampler.}
We queried the unmodified llama.cpp server for the post-sampling probability of the loop token along the repetition path of 12 units (Qwen2.5-1.5B).
For six ordinary sentences, top-$p$ locks between repetitions 5 and 15 and DRY removes every lock.
Six breaker-dense units (short lines, a key--value line, a JSON object, quoted code) lock under DRY at exactly the repetition at which they lock without it.

\paragraph{Free generation.}

\begin{figure}[t]
\centering
\includegraphics[width=0.95\columnwidth]{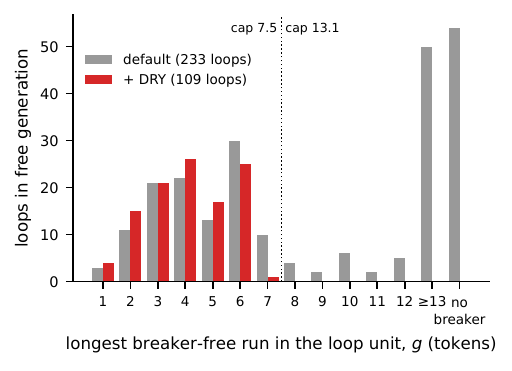}
\caption{Loops found in free generation (four runs pooled), by the longest breaker-free run $g$ of the loop unit. The dotted line separates $D_7=7.5$ from $D_8=13.1$.}
\label{fig:gap}
\end{figure}

Figure~\ref{fig:gap} pools the loops found in four free-generation runs (Qwen3-4B on both prompt sets, Qwen3.6-35B, Llama-3.1-8B; 1{,}780 generations per setting in total).
With the default sampler there are 233 loops, of which 110 have $g\le7$ and 123 have $g\ge8$ or no breaker.
With DRY added there are 109 loops, and all 109 have $g\le7$.
No loop with $g\ge8$ survives DRY in any run (Appendix~\ref{app:details}); 62 of the 109 come from three structured prompts.
On Qwen3-8B, Gemma-2-9B and Phi-4-mini, 110 of the 215 default loops have $g\ge8$ or no breaker, and all 90 loops under DRY have $g\le7$ (Table~\ref{tab:more}).

\subsection{Closing the blind spot}

We implemented the count term as a patch to the llama.cpp DRY sampler, disabled by default; with it disabled, outputs are identical to the unpatched build.
We evaluate it in both deployments: added to DRY (DRY-C), and in place of DRY's length term (count only).

\begin{figure}[t]
\centering
\includegraphics[width=0.9\columnwidth]{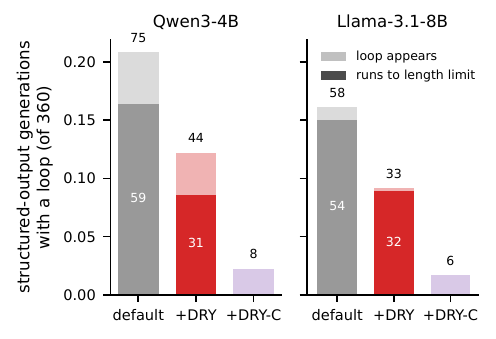}
\caption{Structured-output prompts (360 generations per bar). Light: a loop appears; dark: the loop continues to the length limit. Numbers are counts.}
\label{fig:fix}
\end{figure}

\paragraph{Teacher-forced.}
With DRY-C ($b_c\in\{1.75,3\}$) no unit stays locked: not in the dose--response experiment at any $g$ on any model, not among the units of Table~\ref{tab:lock}, and not among the six breaker-dense units in the real sampler. Count only leaves none of the 1,650 units of Table~\ref{tab:lock} locked (33 lock for a while and unlock later).

\paragraph{Added to DRY.}
Figure~\ref{fig:fix} shows the effect on structured prompts.
On Qwen3-4B, loops that run to the length limit fall from 59 (default) and 31 (DRY) to 0; on Llama-3.1-8B from 54 and 32 to 0.
Loops still \emph{appear} in 8 and 6 generations but end after a median of 6 and 5 repetitions: once the count term starts, each further repetition multiplies its lower bound by $b_c$ (Proposition~\ref{prop:count}).
The share of generated tokens that lie inside loops drops from 12.4\% to 6.9\% to 0.2\% on Qwen3-4B and from 16.3\% to 9.5\% to 0.1\% on Llama-3.1-8B.
The three further models agree: loops at the limit go from 86, 119 and 1 (default) to 34, 53 and 0 (DRY) to 0 on Qwen3-8B, Phi-4-mini and Gemma-2-9B, with coverage within 1.3 points of DRY (Table~\ref{tab:more}).

\begin{table}[t]
\centering\footnotesize
\setlength{\tabcolsep}{3.2pt}
\begin{tabular}{@{}llrrrrr@{}}
\toprule
 & Sampler & Ans. & $\ge$90\% & Cov. & Dup. & JSON \\
\midrule
Qwen3 & default & 62.2 & 15.0 & 34.0 & 28.1 & 26.7 \\
4B & + DRY & 68.3 & 20.0 & 40.5 & 18.5 & 31.1 \\
 & + DRY-C & 65.6 & 20.3 & 38.2 & 15.0 & 31.1 \\
 & pure & 69.2 & 21.4 & 40.1 & 18.5 & 24.4 \\
\midrule
Llama & default & 100 & 52.8 & 66.1 & 23.0 & 13.3 \\
3.1-8B & + DRY & 100 & 58.9 & 71.5 & 12.9 & 26.7 \\
 & + DRY-C & 100 & 61.7 & 73.1 & 8.7 & 22.2 \\
 & pure & 100 & 23.1 & 46.7 & 8.9 & 8.9 \\
\bottomrule
\end{tabular}
\caption{Task completion on structured prompts (\%, 360 generations per row). Ans.: an answer is produced within the limit (for Qwen3-4B, the reasoning block ends); $\ge$90\%: at least 90\% of the requested number of distinct items; Cov.: mean fraction of requested items produced; Dup.: duplicated items among those produced; JSON: outputs that parse, on the three JSON prompts (45 generations per cell). Pure: temperature 1, no truncation.}
\label{tab:quality}
\end{table}

\paragraph{What DRY-C costs.}
Relative to DRY, DRY-C lowers item duplication on four of the five models and moves mean coverage by at most 2.3 points (Tables~\ref{tab:quality} and~\ref{tab:more}).
On 24 tasks with reference answers, record-level F1 under DRY-C differs from DRY by $+2.3$ points (95\% CI $[-0.9,5.5]$) on Qwen and $-1.9$ ($[-5.8,2.0]$) on Llama; both samplers stay below the default sampler's 31.3\% and 31.5\% (Appendix~\ref{app:struct}).
On GSM8K and MATH-500, DRY-C has no loss relative to DRY, and it generates about 5\% more slowly (Appendix~\ref{app:additional}).
DRY itself lowers accuracy by 2.5 to 14.2 points against the default sampler (Appendix~\ref{app:reason}): its length term penalizes a match from three tokens on, and derivations reuse expressions constantly.

\paragraph{Requested repetition.}
When the user asks for verbatim repetition, DRY and DRY-C both get in the way: completion falls from above 90\% to about 10\% (Table~\ref{tab:legit}).
The count term also fires on repeated format skeletons (Appendix~\ref{app:additional}) and costs extra on blocks of short repeated lines; $c_0=4$ recovers part of it with no loop at the limit, while loops rise from 8 and 6 to 39 and 21 (Table~\ref{tab:c0}).

\paragraph{Replacing DRY's length term.}
Count only keeps the guarantee of Corollary~\ref{cor:count} and drops the length term.
Its GSM8K and MATH-500 accuracy returns to the default level on both models (Appendix~\ref{app:reason}).
On the structured prompts of the five models, 92 of 1,800 generations contain a loop (DRY: 167; DRY-C: 34) and one reaches the limit (Table~\ref{tab:count}).
Coverage is above the default sampler's on all five models and below DRY's on four, by 4.2 to 6.5 points.
On the stricter 16-task copying test it recovers about half of what DRY loses (76 and 62 successes of 160, against 29 and 17 under DRY and 126 and 120 by default), and F1 on the reference tasks is 26.5\% and 26.8\%, 7.9 and 2.0 points above DRY (Appendix~\ref{app:reason}).
The length term thus trades reasoning accuracy and copying for new items, while the count term closes the blind spot with or without it: DRY-C suits itemized output, and count only suits reasoning and copying.

\paragraph{Other remedies.}
Dropping DRY's breakers also removes loops but roughly halves JSON validity.
A multiplicative penalty of 1.1 over 4,096 tokens gives Qwen its best coverage and JSON validity, 52.9\% and 51.1\%, with 22 loops at the limit; adding a multiplicative penalty of 1.05 to DRY-C raises Qwen's coverage to 46.8\% and JSON validity to 51.1\% with no loop at the limit (Appendix~\ref{app:additional}).

\section{Conclusion}

Loop survival is a product of per-step continuation probabilities, set by how strongly the model prefers the loop and by what the sampler does with that preference; a lock is its limiting case.
Truncation can remove the exit outright; a bounded penalty only raises the bar, while a penalty that keeps growing reopens it.
DRY's breakers make it a bounded penalty on exactly the loops that matter most for structured output.
A count term lifts the cap and keeps the breakers.
Whether to keep DRY's length term next to it is a separate choice: the length term brings both the reasoning cost and the gain in new items, so DRY-C suits itemized output and the count term alone suits reasoning and copying.

\par\label{main:end}
\section*{Use of AI Assistance}

Claude (Anthropic) helped the authors draft proofs, code and text; the authors checked all of it and take full responsibility for the content.

\bibliography{refs}

\appendix

\section{Proofs}\label{app:proofs}

\begin{proof}[Proof of Proposition~\ref{prop:prod}]
The event $\{T_k\ge m\}$ says that the next $mL$ tokens equal the tokens of the loop.
Write its probability by the chain rule in generation order.
Each factor is conditioned on no deviation so far, in which case the context is exactly some state $(j,l)$ and the factor is $q^S_{j,l}$.
Grouping the $L$ factors of each copy gives $Q_j$ for $j=k,\dots,k+m-1$.
The events decrease to $\{T_k=\infty\}$, and continuity of probability gives the second formula.
\end{proof}

\begin{lemma}\label{lem:prod}
Let $0\le a_j<1$. Then $\prod_j(1-a_j)>0$ iff $\sum_ja_j<\infty$.
\end{lemma}
\begin{proof}
For $0\le x<1$, $-\ln(1-x)\ge x$; for $0\le x\le\frac12$, $-\ln(1-x)\le2x$ (compare derivatives).
If $\sum a_j<\infty$ then $a_j\le\frac12$ for $j\ge J$ and $\sum_{j\ge J}-\ln(1-a_j)\le2\sum_{j\ge J}a_j<\infty$, so the tail product is positive, and the first $J-1$ factors are positive.
If $\sum a_j=\infty$ then $-\ln\prod_{j\le N}(1-a_j)\ge\sum_{j\le N}a_j\to\infty$.
\end{proof}

\begin{proof}[Proof of Proposition~\ref{prop:pure}]
A temperature-scaled softmax of finite logits gives every token a probability strictly between 0 and 1, so $q^S_{n,l}<1$.
The second claim is Proposition~\ref{prop:prod} with Lemma~\ref{lem:prod} applied to $a_j=1-Q_j$.
For the third, $\PP(T_k\ge m)\le\bar Q^m$ and $\EE[T_k]=\sum_{m\ge1}\PP(T_k\ge m)\le\bar Q/(1-\bar Q)$.
\end{proof}

\begin{proof}[Proof of Theorem~\ref{thm:lock}]
A state is locked iff the retained set is $\{u_l\}$.

(a) Suppose $q\ge\pi>\frac12$. Any other token $v$ has $p(v)\le1-q<q$, so the mass strictly above $v$ is at least $q\ge\pi$ and $v$ is dropped; nothing lies strictly above $u_l$, so $u_l$ is kept.
Conversely suppose $q<\pi$. If $u_l$ is not a most probable token, a most probable token $v_1\ne u_l$ is kept. If $u_l$ is the most probable token, let $v_2$ be most probable among the rest; the mass strictly above $v_2$ is at most $q<\pi$ (if several tokens tie with $v_2$ the mass strictly above them is still only $q$), so $v_2$ is kept.

(b) If $u_l$ is not the most probable token, the most probable token is kept and $\ln q-\ln\tilde q\le0$. Otherwise all other tokens are dropped iff $\tilde q<\beta q$.

(c) By definition. The last sentence of the theorem follows from Proposition~\ref{prop:prod}.
\end{proof}

\begin{lemma}\label{lem:power}
Let $x_1,\dots,x_m\ge0$. If $s\ge1$ then $(\max_ix_i)^s\le\sum_ix_i^s\le(\sum_ix_i)^s$. If $0<s\le1$ then $(\sum_ix_i)^s\le\sum_ix_i^s\le m^{1-s}(\sum_ix_i)^s$.
\end{lemma}
\begin{proof}
The case $A=\sum_ix_i=0$ is trivial; otherwise normalize by $A$ and put $y_i=x_i/A\in[0,1]$. For $s\ge1$, $y_i^s\le y_i$; for $s\le1$, $y_i^s\ge y_i$; summing gives the bounds involving $(\sum_i x_i)^s$. The remaining bound for $s\ge1$ is trivial and for $s\le1$ is Jensen's inequality for the concave map $t\mapsto t^s$.
\end{proof}

\begin{proof}[Proof of Proposition~\ref{prop:temp}]
(A) The retained set is determined by the raw distribution; temperature only reweights tokens inside it, and a singleton has probability 1 regardless.

(B) Let $s=1/\tau\ge1$. After temperature the loop token has probability $q^{(\tau)}=q^s/(q^s+\sum_{v\ne u_l}p(v)^s)$. By Lemma~\ref{lem:power}, $\tilde q^{\,s}\le\sum_{v\ne u_l}p(v)^s\le(1-q)^s$, so
\[
\frac{q^s}{q^s+(1-q)^s}\le q^{(\tau)}\le\frac{q^s}{q^s+\tilde q^{\,s}}.
\]
By Theorem~\ref{thm:lock}(a) the state is locked iff $q^{(\tau)}\ge\pi$. The left side is at least $\pi$ iff $\logit(q)\ge\tau\logit(\pi)$; if the state is locked the right side is at least $\pi$, i.e.\ $(q/\tilde q)^s\ge\pi/(1-\pi)$.
For min-$p$, temperature preserves the order of tokens and raises probability ratios to the power $1/\tau$, so by Theorem~\ref{thm:lock}(b) the state is locked iff $(q/\tilde q)^{1/\tau}>1/\beta$.

For $\tau>1$ and top-$p$, the second part of Lemma~\ref{lem:power} with $m=|V|-1$ gives in the same way
\begin{align*}
&\logit(q)\ge\tau\logit(\pi)+(\tau-1)\ln m\ \Rightarrow\ \text{locked},\\
&\text{locked}\ \Rightarrow\ \logit(q)\ge\tau\logit(\pi).
\end{align*}
\end{proof}

\begin{proof}[Proof of Theorem~\ref{thm:pen}]
The penalized distribution is proportional to $p(v)e^{-d(v)}$, which gives the formula for $q'$.
Since $e^{-d(v)}\le1$, the sum in the denominator is at most $1-q$; keeping only unseen tokens, which have $d(v)=0$, it is at least $\ro$.
As $c\mapsto x/(x+c)$ is decreasing this gives the two-sided bound.
The lower bound has log-odds $\logit(q)-d$ and the upper bound has log-odds $\ln(q/\ro)-d$; Theorem~\ref{thm:lock}(a) applied to $q'$ gives the two implications.
The min-$p$ statement is Theorem~\ref{thm:lock}(b) for the penalized distribution; note that the strongest rival may change after the penalty.
\end{proof}

\begin{proof}[Proof of Corollary~\ref{cor:bounded}]
$q\ge\sigma(\logit(\pi)+D)$ iff $\logit(q)\ge\logit(\pi)+D$, which implies $\logit(q)-d\ge\logit(\pi)$.
\end{proof}

\begin{proof}[Proof of Corollary~\ref{cor:growing}]
By the upper bound of Theorem~\ref{thm:pen}, $q'_{n,l}\le q_{n,l}e^{-d_{n,l}}/\ro_{n,l}=\exp\bigl(-(d_{n,l}-\ln(q_{n,l}/\ro_{n,l}))\bigr)\to0$.
The set retained by top-$p$ has mass at least $\pi$, so after renormalization $q^S_{n,l}\le q'_{n,l}/\pi$ (or $0$ if the loop token is dropped).
Hence $Q_n\le q'_{n,l}/\pi\to0$ and $\PP(T_k=\infty)\le Q_N\to0$ by Proposition~\ref{prop:prod}.
If $d_{n,l}\to\infty$ and $\ro_{n,l}\ge\rho>0$, then $d_{n,l}-\ln(q_{n,l}/\ro_{n,l})\ge d_{n,l}-\ln(1/\rho)\to\infty$.
\end{proof}

\begin{proof}[Proof of Proposition~\ref{prop:dry}]
If $u_l\in B$ the penalty is zero by definition.
Let $u_l\notin B$.
Since $n\ge1$, the context ends with at least one complete copy of $u$, which contains a breaker; the $\kappa$ trailing non-breaker tokens therefore lie in the loop and form a contiguous arc of the circle.
The token $u_l$ extends this arc, giving $\kappa+1$ consecutive non-breaker tokens, so $\kappa+1\le g$.
Thus $M(u_l)\le\kappa\le g-1$.
If $g-1<a$ the penalty is zero; otherwise it is at most $\lambda b^{g-1-a}$ because $b>1$.
For attainment, let $u_l$ be the last token of a longest breaker-free arc, so that $\kappa=g-1$.
For $n\ge2$ the position $j=t-L$ satisfies $x_{j+1}=u_l$ and is preceded by at least one full copy that agrees with the current suffix, so $M_j\ge L>\kappa$; it lies in the window with $j-h+1=\min\{t-L,\,W-L\}\ge g-1$, because $t-L\ge L>g$ when $n\ge2$ (the unit contains a breaker, so $L\ge g+1$) and $W-L\ge g-1$ by assumption, so $M(u_l)=g-1$.
\end{proof}

\begin{proof}[Proof of Corollary~\ref{cor:blind}]
Proposition~\ref{prop:dry} bounds the penalty on the loop token by $D_g$.
DRY may also penalize other candidates, but the sufficient condition of Theorem~\ref{thm:pen} holds for arbitrary $d(v)\ge0$, so Corollary~\ref{cor:bounded} applies.
\end{proof}

\begin{proof}[Proof of Proposition~\ref{prop:count}]
Let $t$ be the last position of the context, so that $u_l$ is to be generated next.
For $k=1,\dots,n$ put $j_k=t-kL$, the same position $k$ copies earlier; then $x_{j_k+1}=u_l$.
From the start of the loop up to $j_k$ there are $(n-k)L+(l-1)$ loop tokens, and by periodicity they agree with the equally long suffix ending at $t$, so $M_{j_k}\ge(n-k)L$.
Therefore $j_k$ is counted whenever $(n-k)L\ge a'$, i.e.\ for $k\le n-\lceil a'/L\rceil$, provided its matching span of $a'$ tokens lies within the last $W$ tokens, i.e.\ $kL+a'\le W$, i.e.\ $k\le N_W$.
Hence $c(u_l)\ge\min(n-\lceil a'/L\rceil,N_W)$, and the bound follows because $b_c>1$.
\end{proof}

\begin{proof}[Proof of Corollary~\ref{cor:count}]
Under top-$p$ alone, the necessary condition of Theorem~\ref{thm:pen} gives $d(u_l)\le\ln(q/\ro)-\logit(\pi)\le\Delta$ in a locked state.
When top-$k$ comes first, call $p(v)e^{-d(v)}$ the weight of $v$; unseen tokens keep their raw probability as weight. A locked state retains $u_l$ and $k-1\ge1$ other tokens. If the unseen token of probability $r^{*}$ is retained, the other retained weights sum to at least $r^{*}$; if it is not, every retained token has weight at least $r^{*}$, again giving at least $r^{*}$ for the others. Top-$p$ then keeps $u_l$ alone only if $qe^{-d}\ge\pi(qe^{-d}+r^{*})$, that is, $d(u_l)\le\ln(q/r^{*})-\logit(\pi)\le\Delta$.
Let $m=\min(n-\lceil a'/L\rceil,N_W)$ and suppose $m>c_0+\delta$. Then $m\ge c_0$, so Proposition~\ref{prop:count} gives $d(u_l)\ge\lambda b_c^{\,m-c_0}>\lambda b_c^{\delta}=\max\{\Delta,\lambda\}\ge\Delta$, and the state is not locked.
If $N_W>c_0+\delta$, then $m>c_0+\delta$ holds exactly when $n>\lceil a'/L\rceil+c_0+\delta$.
\end{proof}

\section{Verification Details}\label{app:verify}

\paragraph{Lock criteria.}
For Qwen2.5-1.5B, Qwen2.5-7B, Qwen3-4B and Llama-3.2-3B we compared, at every position of every unit, the lock status computed exactly from the full distribution with the status predicted from $q$ and $\tilde q$ alone.
For top-$p\in\{0.9,0.95\}$ and min-$p\in\{0.1,0.05\}$ the two agree at all positions (about 160{,}000 per model).
A Monte Carlo check with the \texttt{transformers} sampler reproduces the survival formula ($0.789\pm0.013$ sampled against $0.786$ computed).
The smallest escape probability $1-q$ observed under pure sampling ranges from $1.9\times10^{-13}$ to $3\times10^{-6}$ across models and is always positive; after 200 repetitions the median per-repetition escape probability $1-Q_n$ ranges from $10^{-1.0}$ (Llama-3.2-3B) to $10^{-3.8}$ (Qwen3-8B) across the eleven models.

\paragraph{Temperature.}
For $\tau=0.7$ before top-$p$ 0.9, the sufficient and necessary conditions of Proposition~\ref{prop:temp}(B) have no violations, and the sufficient condition covers 99.6\%--99.9\% of locked positions.
The median onset of the final lock comes earlier on all eleven models, from repetitions 5--110 to 3--38.

\paragraph{Penalties.}
For presence penalties 1 and 2, the sufficient condition of Theorem~\ref{thm:pen} has no violations and covers 99.2\%--100\% of locked positions.
For these and for full-context frequency penalties 0.1 and 0.5, the necessary condition with the measured $\ro$ has no violations, and some locked positions meet it with zero slack (to two decimals); replacing $\ro$ by $1-q$ produces tens to thousands of violations, so the distinction matters.

\section{Simulation versus the llama.cpp Sampler}\label{app:sim}

The teacher-forced DRY results rely on our implementation of the definition in Section~\ref{sec:dry}.
To compare samplers rather than weights, we give our implementation and llama.cpp (commit \texttt{03aa006}) the same logits.
For 12 units and 40 repetitions (5{,}072 positions, 5{,}049 with values in every request), one request returns the server's probabilities of the 400 most likely tokens before any penalty; we apply our DRY and top-$p$ 0.9 to them, treating the mass beyond these 400 tokens as unpenalized.
A second request returns the server's own probabilities after DRY and top-$p$ 0.9.
Lock status agrees at all 5{,}049 positions (2{,}313 locked), and the loop token's probability differs by at most $0.019$.
The count term, checked the same way against the patched sampler, also agrees on lock status everywhere (311 locked), with differences of at most $0.029$.
Candidate sets differ mainly where the penalty has removed the loop token in both computations and the remaining mass lies near the end of the 400 returned tokens; where the loop token is kept, they differ at 17 (DRY) and 33 (DRY-C) positions, by at most $0.036$ and $0.043$.
We disable prompt caching here, since cached prefixes shift tail log-probabilities by up to $0.14$.
With the patch disabled, the patched and unpatched builds return identical probabilities at all 5{,}072 positions for top-$p$ and at all positions where both return a value for DRY.

\section{Experimental Details}\label{app:details}

\paragraph{Teacher-forced runs.}
In the window reruns, Qwen2.5-1.5B uses \texttt{float32}, both Gemma models use \texttt{bfloat16}, and the other models use \texttt{float16}. Softmax and sampler calculations use \texttt{float64}.
Contexts are limited to 12{,}288 tokens, to 8{,}192 tokens (their native context) for the Gemma models, and to 4{,}000 tokens for Phi-4-mini, whose outputs degrade abruptly beyond 4{,}096 tokens under chunked evaluation.
For Gemma~2 we apply the model's final logit soft cap, $z\mapsto30\tanh(z/30)$, before any sampler step.
Reinforcement curves use the inline variants of the corpus sentences whose tokenization is stable under repetition (277--300 per model); lock fractions and the DRY comparison use an evenly spaced subset of 150 units; the dose--response experiment uses 40 base sentences per model that contain no breaker characters or semicolons.
A unit is counted as finally locked if its last ten measured repetitions are entirely locked (Table~\ref{tab:lock} and the 60-repetition dose experiment). Ever-locked means at least one measured repetition is entirely locked.
In the third run (count only), Gemma-2-9B has 141 units locked under top-$p$ instead of 142; the difference comes from the evaluation chunk size (32 versus 256 tokens) in \texttt{bfloat16} for units near the threshold.

\paragraph{Free generation.}
Quantizations: Q8\_0 for Qwen3-4B and Llama-3.1-8B, Q4\_K\_M for Qwen3.6-35B-A3B.
Prompt set~1: 25 samples per prompt and setting for Qwen3-4B, 10 for the 35B model. Prompt set~2: 15 samples per prompt and setting.
DRY uses $\lambda=0.8$, $b=1.75$, $a=2$, the default breakers, and a range of 4{,}096 tokens.
Table~\ref{tab:runs} lists loop counts per run, and Table~\ref{tab:allset} gives every sampler setting run on prompt set~1.
The windowed multiplicative penalty (rep.\ 1.1) yields fewer loops there than DRY, which we attribute to its also penalizing breaker tokens; in teacher forcing it leaves 98\% of Qwen3-4B's units locked (Table~\ref{tab:lock}).

\begin{table}[h]
\centering\footnotesize
\setlength{\tabcolsep}{2pt}
\begin{tabular}{@{}lrrrrr@{}}
\toprule
 & default & + DRY & + pres.\ 1 & + rep.\ 1.1 & pure \\
\midrule
Qwen3-4B & 78 & 28 & 44 & 24 & 37 \\
\quad of which $g\le7$ & 30 & 28 & 29 & 19 & 26 \\
Qwen3.6-35B & 9 & 4 & 1 & 0 & 1 \\
\quad of which $g\le7$ & 3 & 4 & 0 & 0 & 0 \\
\bottomrule
\end{tabular}
\caption{Loops on prompt set~1 under every sampler setting (500 generations per cell for Qwen3-4B, 200 for the 35B model). Pure: temperature 1, no truncation.}
\label{tab:allset}
\end{table}
DRY-C uses $a'=24$, $c_0=2$, $b_c=3$; with $b_c=1.75$ on Qwen3-4B (prompt set~2), 25 structured generations contain a loop instead of 8, and again none runs to the length limit.

\begin{table}[h]
\centering\footnotesize
\setlength{\tabcolsep}{2.5pt}
\begin{tabular}{@{}llrrr@{}}
\toprule
Run & Sampler & loops & $g\le7$ & $g\ge8$ \\
\midrule
Qwen3-4B, set 1 & default & 78 & 30 & 48 \\
 & + DRY & 28 & 28 & 0 \\
Qwen3-4B, set 2 & default & 75 & 45 & 30 \\
 & + DRY & 44 & 44 & 0 \\
Qwen3.6-35B, set 1 & default & 9 & 3 & 6 \\
 & + DRY & 4 & 4 & 0 \\
Llama-3.1-8B, set 2 & default & 71 & 32 & 39 \\
 & + DRY & 33 & 33 & 0 \\
\bottomrule
\end{tabular}
\caption{Loops per free-generation run. The last column includes units without any breaker. Prompt set~2 includes both structured and prose prompts; on Llama, 58 of the 71 default loops and all 33 DRY loops occur on the structured subset used in Table~\ref{tab:baselines}.}
\label{tab:runs}
\end{table}
Within the blind spot the distribution shifts: at $g=7$, where the cap $D_7=7.5$ is close to typical log-odds, loops drop from 10 to 1, whereas loops with $g\le5$ rise from 70 to 83: generations that DRY steers away from a long-run loop can fall into a shorter-run one.

\paragraph{Sensitivity to the loop criterion.}
Table~\ref{tab:crit} repeats the count of Figure~\ref{fig:gap} under looser and stricter loop criteria.
The number of loops changes, but under every criterion no loop that appears with DRY enabled has $g\ge8$.
The same holds when every loop interval in a generation is counted, not only the first: in the 289 generations on prompt set~2 that contain a loop under some configuration keeping DRY's length term (DRY-C and the baselines of Appendix~\ref{app:additional} included), no interval has $g\ge8$ or lacks a breaker.

\begin{table}[h]
\centering\footnotesize
\setlength{\tabcolsep}{2.5pt}
\begin{tabular}{@{}lrrrrrr@{}}
\toprule
 & \multicolumn{3}{c}{default} & \multicolumn{3}{c}{default + DRY} \\
\cmidrule(lr){2-4}\cmidrule(l){5-7}
Criterion & all & $g\le7$ & $g\ge8$ & all & $g\le7$ & $g\ge8$ \\
\midrule
$\ge$3 reps, $\ge$30 tok. & 309 & 124 & 185 & 129 & 129 & 0 \\
$\ge$4 reps, $\ge$40 tok. & 262 & 120 & 142 & 116 & 116 & 0 \\
$\ge$5 reps, $\ge$60 tok. & 233 & 110 & 123 & 109 & 109 & 0 \\
$\ge$8 reps, $\ge$100 tok. & 184 & 95 & 89 & 99 & 99 & 0 \\
$\ge$10 reps, $\ge$150 tok. & 169 & 91 & 78 & 90 & 90 & 0 \\
\bottomrule
\end{tabular}
\caption{Loops in the four pooled free-generation runs under different detection criteria (minimum consecutive repetitions and minimum total length). The third row is the criterion used in the paper.}
\label{tab:crit}
\end{table}

\paragraph{Requested repetition.}
Twelve prompts ask for verbatim repetition of a text given in the prompt: four ask for a sentence to be copied 5 or 8 times, two for a paragraph to be copied 3 times, two for a four-line chorus after each of 4 verses, and four for a fixed line in each of 5--8 items (notes, functions, SQL statements); half are in English and half in Chinese.
A generation succeeds if the given text occurs the required number of times in the answer, ignoring differences in whitespace.
Table~\ref{tab:legit} reports this 12-task experiment, with 10 generations per task for default and DRY (120 per row) and three batches of 10 per task for each DRY-C setting (360 per row). A stricter 16-task version is reported in Appendix~\ref{app:additional}.
On Qwen3-4B the chorus tasks, whose lines are short and newline-terminated, are the only ones that still succeed under DRY; on Llama-3.1-8B no chorus generation succeeds.

\begin{table}[h]
\centering\footnotesize
\setlength{\tabcolsep}{1.5pt}
\begin{tabular}{@{}llrrrrr@{}}
\toprule
Model & Sampler & sent. & para. & chorus & templ. & all \\
\midrule
Llama-3.1-8B & default & 100 & 100 & 70 & 100 & 95 \\
 & + DRY & 22 & 0 & 0 & 8 & 10 \\
 & + DRY-C & 22 & 0 & 0 & 8 & 10 \\
 & + DRY-C, $c_0$=6 & 23 & 0 & 0 & 7 & 10 \\
Qwen3-4B & default & 100 & 100 & 80 & 82 & 91 \\
 & + DRY & 0 & 0 & 75 & 0 & 12 \\
 & + DRY-C & 0 & 0 & 75 & 0 & 12 \\
 & + DRY-C, $c_0$=6 & 1 & 0 & 75 & 0 & 13 \\
\bottomrule
\end{tabular}
\caption{Tasks that request verbatim repetition: generations (\%) containing the required number of copies, ignoring whitespace.}
\label{tab:legit}
\end{table}

\paragraph{Structured versus prose.}
With the default sampler, 20.8\% (Qwen3-4B) and 16.1\% (Llama-3.1-8B) of structured generations on prompt set~2 contain a loop, against 0\% and 7.2\% for prose; the prose loops of Llama all have $g\ge8$ or no breaker, and DRY removes all of them.

\paragraph{Further models.}
Qwen3-8B (Q8\_0, Qwen3 sampler), Gemma-2-9B (Q6\_K) and Phi-4-mini (Q8\_0), the last two with $\tau=0.6$ and top-$p$ 0.95, ran the 24 structured prompts of set~2 with 15 samples per prompt and setting (Table~\ref{tab:more}).
Under DRY all 90 loops have $g\le7$ in every interval, and DRY-C has no loop at the limit on any of the three.

\begin{table}[h]
\centering\scriptsize
\setlength{\tabcolsep}{1.9pt}
\begin{tabular}{@{}llrrrrrrrr@{}}
\toprule
 & Sampler & Loops & Limit & $g\le7$ & $g\ge8$ & Tok. & Cov. & Dup. & JSON \\
\midrule
Qwen3 & default & 90 & 86 & 35 & 55 & 17.6 & 37.6 & 23.6 & 35.6 \\
8B & + DRY & 34 & 34 & 34 & 0 & 6.7 & 47.2 & 13.7 & 40.0 \\
 & + DRY-C & 7 & 0 & 7 & 0 & 0.1 & 46.8 & 12.5 & 44.4 \\
\midrule
Gemma-2 & default & 4 & 1 & 4 & 0 & 0.8 & 44.6 & 5.0 & 46.7 \\
9B & + DRY & 1 & 0 & 1 & 0 & 0.1 & 45.6 & 2.1 & 55.6 \\
 & + DRY-C & 0 & 0 & 0 & 0 & 0.0 & 44.3 & 2.4 & 64.4 \\
\midrule
Phi-4 & default & 121 & 119 & 66 & 55 & 43.6 & 33.2 & 44.3 & 33.3 \\
mini & + DRY & 55 & 53 & 55 & 0 & 19.5 & 44.5 & 26.3 & 33.3 \\
 & + DRY-C & 13 & 0 & 13 & 0 & 0.5 & 43.6 & 24.3 & 48.9 \\
\bottomrule
\end{tabular}
\caption{Three further models, 360 structured generations per row. Limit: a loop runs to the length limit; $g\ge8$: some interval has $g\ge8$ or no breaker; Tok.: loop-token share (\%); Cov., Dup., JSON as in Table~\ref{tab:quality} (\%).}
\label{tab:more}
\end{table}

\paragraph{Larger models.}
The 35B model loops far less than the 4B model under the same sampler and prompts (4.5\% of generations against 15.6\%), in agreement with \citet{pipis2025wait}; the four loops that survive DRY all have $g\le6$.

\begin{figure}[h]
\centering
\includegraphics[width=0.95\columnwidth]{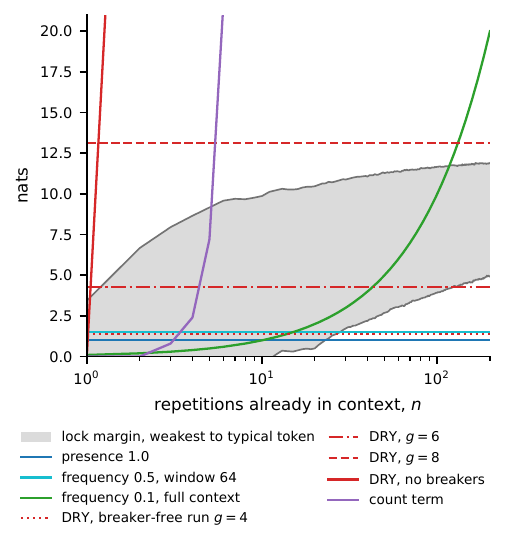}
\caption{Penalty trajectories (analytic, unit length 24, unlimited look-back for growing curves; ``count term'' is the DRY-C count component) against the lock margin $\logit(q)-\logit(0.9)$ of Qwen3-4B (band: weakest to typical token of the median sentence). A penalty that stays below the band cannot unlock.}
\label{fig:pen}
\end{figure}

\paragraph{Penalty trajectories.}
Figure~\ref{fig:pen} shows each penalty on the loop token as a function of $n$, contrasting fixed caps with linear and exponential lower-bound trajectories. The curve labeled ``count term'' shows the count component alone, not the maximum of the DRY and count penalties.

\paragraph{Task completion.}
For each structured prompt we extract items from the answer (for reasoning models, the text after the reasoning block) with a prompt-specific pattern, and compare the number of distinct items with the number requested.
Generations that reach the length limit before producing an answer count as zero items.

\section{Additional Baselines and Robustness}\label{app:additional}
\setlength{\floatsep}{8pt}\setlength{\dblfloatsep}{8pt}

\paragraph{Additional interventions.}
Antislop suppresses unwanted strings by backtracking \citep{v5antislop}; Local Temperature Beam Search uses deterministic calibration \citep{v5ltbs}; Variance-Calibrated Modulation adjusts decoding to address concentration in the distribution head \citep{v5vcm}; and XTC excludes top choices \citep{v5xtc}.
e-CUSUM studies reasoning monitors \citep{v5ecusum}, ``One Neuron'' explores internal-component edits \citep{v5neuron}, ``Feedback That Backfires'' examines repeated failed actions after feedback \citep{v5feedback}, ``Word Salad Chopper'' detects unproductive reasoning from hidden states \citep{v5salad}, ``Circular Reasoning'' ties reasoning loops to self-reinforcing attention and flags them early with CUSUM \citep{duan2026circular}, and LoopGuard prunes the repeated tail from the key--value cache \citep{xu2026loopguard}.

\paragraph{Additional baselines.}
Table~\ref{tab:baselines} uses the same 24 structured prompts, 15 samples per prompt and setting, and 3,000-token generation limit as Table~\ref{tab:quality}.
The two token penalties with $W=4096$ look back 4,096 tokens, which covers the whole generation, so the frequency penalty grows throughout.
The bounded alternatives (rep.\ 1.1, DRY + rep.\ 1.05, newline-only breakers) still produce loops that run to the generation limit. Frequency 0.1 leaves few such loops (6 and 7), but on Llama lowers mean item coverage from 66.1\% (default) to 43.3\%, with only 7.8\% of generations reaching 90\% of the requested distinct items.
Removing every breaker eliminates detected loops in both models while reducing JSON validity relative to DRY (31.1\% to 17.8\% for Qwen; 26.7\% to 11.1\% for Llama).
DRY-C again has no loop reaching the limit; mean coverage is 38.2\% against 40.5\% on Qwen and 73.1\% against 71.5\% on Llama.
Adding rep.\ 1.05 to DRY-C on Qwen yields 3 loops, none to the limit, 46.8\% coverage and 51.1\% JSON validity; the corresponding DRY + rep.\ 1.05 baseline has 13 run-to-limit loops and 47.4\% coverage.
On Llama the same combination yields no loop, 72.6\% coverage and 24.4\% JSON validity, against 6 loops, 73.1\% and 22.2\% for DRY-C alone and 5 run-to-limit loops for DRY + rep.\ 1.05.

\paragraph{Ablations and cost.}
The count term does the work by itself: with DRY's length term removed, 16 structured generations on Qwen3-4B and 8 on Llama contain a loop and none reaches the limit (Table~\ref{tab:baselines}, Appendix~\ref{app:reason}).
A linear count term $\lambda(c-c_0+1)$ also keeps growing until the window is saturated; on Qwen3-4B it leaves 23 generations with a loop and one at the limit, so the exponential term ends loops sooner.
Replacing the count term by a hard block, which excludes every continuation with $c\ge c_0$, leaves no detected loop in either model; coverage is 39.5\% on Qwen and 72.0\% on Llama, and JSON validity 20.0\% on both.
On Qwen3-4B (8-bit GGUF, one RTX 3090, eight prompts of 3{,}000 tokens), llama.cpp generates 100.4 tokens/s with DRY-C against 105.6 with DRY, a 5\% slowdown.

\begin{table*}[t]
\centering\footnotesize\renewcommand{\arraystretch}{0.92}
\begin{tabular}{@{}llrrrrr@{}}
\toprule
Model & Sampler & Loops & To limit & Loop tokens (\%) & Coverage (\%) & JSON (\%) \\
\midrule
Qwen3-4B & default & 75 & 59 & 12.4 & 34.0 & 26.7 \\
 & DRY & 44 & 31 & 6.9 & 40.5 & 31.1 \\
 & DRY-C & 8 & 0 & 0.2 & 38.2 & 31.1 \\
 & frequency 0.1, $W=4096$ & 19 & 6 & 0.8 & 39.6 & 15.6 \\
 & rep.\ 1.1, $W=4096$ & 31 & 22 & 4.2 & 52.9 & 51.1 \\
 & DRY + rep.\ 1.05 & 20 & 13 & 2.9 & 47.4 & 42.2 \\
 & DRY, newline breakers only & 31 & 29 & 4.9 & 43.3 & 15.6 \\
 & DRY, no breakers & 0 & 0 & 0.0 & 42.7 & 17.8 \\
 & DRY-C + rep.\ 1.05 & 3 & 0 & 0.3 & 46.8 & 51.1 \\
 & count term only & 16 & 0 & 0.2 & 36.3 & 44.4 \\
 & DRY + linear count & 23 & 1 & 1.0 & 38.0 & 37.8 \\
 & DRY + hard count block & 0 & 0 & 0.0 & 39.5 & 20.0 \\
 & pure, $\tau=0.6$ & 72 & 53 & 11.5 & 30.4 & 31.1 \\
\midrule
Llama-3.1-8B & default & 58 & 54 & 16.3 & 66.1 & 13.3 \\
 & DRY & 33 & 32 & 9.5 & 71.5 & 26.7 \\
 & DRY-C & 6 & 0 & 0.1 & 73.1 & 22.2 \\
 & frequency 0.1, $W=4096$ & 85 & 7 & 8.1 & 43.3 & 4.4 \\
 & rep.\ 1.1, $W=4096$ & 7 & 4 & 1.7 & 63.3 & 35.6 \\
 & DRY + rep.\ 1.05 & 6 & 5 & 1.8 & 67.0 & 33.3 \\
 & DRY, newline breakers only & 14 & 14 & 4.9 & 70.1 & 33.3 \\
 & DRY, no breakers & 0 & 0 & 0.0 & 60.7 & 11.1 \\
 & DRY-C + rep.\ 1.05 & 0 & 0 & 0.0 & 72.6 & 24.4 \\
 & count term only & 8 & 0 & 0.2 & 66.6 & 26.7 \\
 & DRY + hard count block & 0 & 0 & 0.0 & 72.0 & 20.0 \\
 & pure, $\tau=0.6$ & 46 & 40 & 12.3 & 65.3 & 31.1 \\
\bottomrule
\end{tabular}
\caption{Additional baselines, 360 structured generations per row. Loops: generations with a detected loop; to limit: generations in which some loop runs to the length limit. Loop tokens: fraction of all generated tokens inside any detected loop; coverage and JSON use the definitions of Table~\ref{tab:quality} (45 JSON generations per row). All penalty variants are added to the model's default sampler. Count term only: DRY-C without DRY's length term; linear count: $\lambda(c-c_0+1)$ in place of the exponential count term; hard count block: every continuation with $c\ge c_0$ is excluded; pure, $\tau=0.6$: temperature only, no truncation.}
\label{tab:baselines}
\end{table*}

\paragraph{Count-parameter sensitivity.}
Table~\ref{tab:c0} varies $c_0$ with $a'=24$ and $b_c=3$.
Through $c_0=4$ no loop reaches the limit on either model; at $c_0=6$ and $8$ a few do.
Strict success on the block tasks climbs toward DRY's level as $c_0$ grows.
Varying $a'$ with $c_0=2$ on Qwen, $a'=12$ gives 4 loops (0.04\% of tokens), 39.1\% coverage and 28.9\% JSON validity, and $a'=48$ gives 33 loops (1.19\%), 38.1\% and 37.8\%, again none at the limit; the larger $a'$ activates later (Proposition~\ref{prop:count}), so more loops meet the detector criterion before ending.

\begin{table}[h]
\centering\footnotesize
\setlength{\tabcolsep}{3pt}
\begin{tabular}{@{}llrrrrrr@{}}
\toprule
 & $c_0$ & Loops & Limit & Tok. & Cov. & JSON & Block \\
\midrule
Qwen3 & DRY & 44 & 31 & 6.9 & 40.5 & 31.1 & 12 \\
4B & 2 & 8 & 0 & 0.2 & 38.2 & 31.1 & 4 \\
 & 3 & 22 & 0 & 0.5 & 38.6 & 37.8 & 5 \\
 & 4 & 39 & 0 & 1.8 & 37.8 & 28.9 & 10 \\
 & 6 & 40 & 1 & 1.9 & 37.8 & 33.3 & 11 \\
 & 8 & 45 & 3 & 2.8 & 35.4 & 28.9 & 12 \\
\midrule
Llama & DRY & 33 & 32 & 9.5 & 71.5 & 26.7 & 17 \\
3.1-8B & 2 & 6 & 0 & 0.1 & 73.1 & 22.2 & 10 \\
 & 3 & 10 & 0 & 0.5 & 71.2 & 35.6 & 9 \\
 & 4 & 21 & 0 & 2.5 & 71.9 & 17.8 & 13 \\
 & 6 & 33 & 2 & 2.8 & 68.9 & 28.9 & 16 \\
 & 8 & 32 & 7 & 4.0 & 72.2 & 26.7 & 15 \\
\bottomrule
\end{tabular}
\caption{DRY-C at different $c_0$ on 360 structured generations per row. Block: strict successes of 40 on the four block tasks of Table~\ref{tab:strict}. Other columns as in Table~\ref{tab:more}.}
\label{tab:c0}
\end{table}

\paragraph{Windowed lock status and finite survival.}
Under the full-context frequency penalty locks still form transiently, as late as repetition 86 (Qwen3-8B), before the linear term overtakes them.
The second run, with a 4,096-token look-back for DRY and DRY-C, covers all eleven models with 150 units each (Phi-4 again capped at 4,000 tokens).
Top-$p$ is ever locked on 1,612/1,650 units and locked through the last ten repetitions on 1,568/1,650. DRY has 25 ever-locked and 21 finally locked units, all final locks with $5\le g\le7$; DRY-C ($b_c=3$) has 3 ever-locked units and no final locks.
In Table~\ref{tab:window}, $k_3$ is the probability, starting with three copies in context, of following the loop to that unit's measured endpoint, averaged over units.

\begin{table*}[t]
\centering\footnotesize\renewcommand{\arraystretch}{0.92}
\begin{tabular}{@{}llrrrrrr@{}}
\toprule
 & & \multicolumn{2}{c}{top-$p$} & \multicolumn{2}{c}{DRY} & \multicolumn{2}{c}{DRY-C} \\
Model & $N$ & ever/final & mean $k_3$ & ever/final & mean $k_3$ & ever/final & mean $k_3$ \\
\midrule
Gemma-2-2B & 150 & 125/106 & 0.00262 & 1/1 & 1.76e-30 & 0/0 & 0 \\
Gemma-2-9B & 150 & 149/142 & 0.00124 & 2/2 & 1.34e-12 & 0/0 & 0 \\
Llama-3.1-8B & 150 & 150/150 & 0.113 & 1/1 & 9.57e-06 & 0/0 & 0 \\
Llama-3.2-3B & 150 & 143/142 & 0.147 & 1/1 & 1.38e-06 & 0/0 & 0 \\
Phi-4-mini & 150 & 150/149 & 0.472 & 0/0 & 6.83e-35 & 0/0 & 0 \\
Qwen2.5-1.5B & 150 & 150/150 & 0.130 & 3/3 & 0.000435 & 0/0 & 0 \\
Qwen2.5-7B & 150 & 150/149 & 0.346 & 3/3 & 0.00173 & 0/0 & 0 \\
Qwen3-1.7B & 150 & 150/149 & 0.272 & 4/3 & 0.0061 & 0/0 & 0 \\
Qwen3-4B & 150 & 150/150 & 0.373 & 4/3 & 0.00839 & 1/0 & 0 \\
Qwen3-8B & 150 & 150/149 & 0.700 & 4/3 & 0.0146 & 2/0 & 0 \\
R1-Distill-7B & 150 & 145/132 & 0.000902 & 2/1 & 2.99e-08 & 0/0 & 0 \\
\bottomrule
\end{tabular}
\caption{Window $W=4096$: ever-locked/finally-locked counts and mean finite-horizon $k_3$. Final: last ten repetitions all locked. DRY-C uses $b_c=3$.}
\label{tab:window}
\end{table*}

\paragraph{Actual breaker gaps.}
Of 3,080 colon units, 2,839 satisfy the top-$p$ last-ten lock criterion. Retokenization changes the actual gap on 51 units overall, including 47 of these 2,839.
Table~\ref{tab:actualgap} contrasts grouping by the intended insertion interval with grouping by the recorded actual token gap.
The transition persists: at actual $g=6$, 250/408 remain locked under DRY; at $g=8$, 1/411 do.
The semicolon control has 0/2,843 remaining locks; its denominator counts the control units (out of 3,080) that lock under top-$p$ alone.

\begin{table}[t]
\centering\small
\begin{tabular}{@{}rrr@{}}
\toprule
$g$ & Nominal interval & Actual gap \\
\midrule
2 & 387/387 & 379/379 \\
3 & 392/398 & 398/404 \\
4 & 385/410 & 385/409 \\
5 & -- & 1/2 \\
6 & 249/407 & 250/408 \\
8 & 1/415 & 1/411 \\
9 & -- & 0/3 \\
10 & -- & 0/1 \\
11 & -- & 0/2 \\
12 & 0/412 & 0/410 \\
23 & -- & 0/29 \\
24 & 0/410 & 0/381 \\
\bottomrule
\end{tabular}
\caption{DRY-locked / top-$p$-locked colon units, pooled over eleven models, under nominal and actual gap grouping.}
\label{tab:actualgap}
\end{table}

\paragraph{Stricter requested-copying evaluation.}
This experiment adds four tasks that repeat blocks of short lines to the twelve tasks of Appendix~\ref{app:details}. Each setting has 16 tasks with 10 samples each (160 generations per model and setting).
We report both whitespace-normalized substring counts and strict counts that normalize only spaces/tabs immediately before a newline, preserving internal spaces and line structure (Table~\ref{tab:strict}).
The strict metric shows additional copying loss from DRY-C: overall success falls from DRY's 10.6\% to 6.3\% on Llama and from 18.1\% to 11.2\% on Qwen at $c_0=2$.
On the four block tasks (40 generations), strict success is 42.5\% to 25.0\% on Llama and 30.0\% to 10.0\% on Qwen. Raising $c_0$ to 6 brings these to 40.0\% and 27.5\%, respectively.

\begin{table}[t]
\centering\small
\begin{tabular}{@{}lrrrr@{}}
\toprule
 & \multicolumn{2}{c}{Llama} & \multicolumn{2}{c}{Qwen} \\
Sampler & norm. & strict & norm. & strict \\
\midrule
default & 127 & 120 & 127 & 126 \\
DRY & 27 & 17 & 31 & 29 \\
DRY + rep.\ 1.05 & 28 & 12 & 35 & 31 \\
DRY, newline only & 32 & 18 & 26 & 24 \\
DRY-C, $c_0=2$ & 26 & 10 & 18 & 18 \\
DRY-C, $c_0=6$ & 30 & 16 & 28 & 22 \\
\bottomrule
\end{tabular}
\caption{Requested repetition, 16-task experiment: successful generations out of 160 in every cell. Norm.: whitespace-normalized; strict: only trailing line whitespace normalized.}
\label{tab:strict}
\end{table}

\paragraph{Count activation outside detected loops.}
Table~\ref{tab:fpgrid} counts teacher-forced activations on default-sampler generations with no loop under our detector, using exact $(a'+1)$-token tuples and a 4,096-token window (all sequences are at most 3,000 tokens).
At $(a',c_0)=(24,2)$, at least one continuation would be penalized in 108/285 (37.9\%) Qwen structured outputs and 57/302 (18.9\%) Llama structured outputs; affected-position shares are 3.0\% and 5.7\%.
For prose the corresponding generation shares are 9/180 (5.0\%) and 50/167 (29.9\%).
Raising either threshold reduces them.

\begin{table}[t]
\centering\small
\begin{tabular}{@{}rrrrr@{}}
\toprule
 & & \multicolumn{2}{c}{Structured} & Prose \\
$a'$ & $c_0$ & Qwen & Llama & Qwen / Llama \\
\midrule
12 & 2 & 68.1 & 53.0 & 14.4 / 49.7 \\
12 & 4 & 42.8 & 42.7 & 2.8 / 19.8 \\
24 & 2 & 37.9 & 18.9 & 5.0 / 29.9 \\
24 & 4 & 14.4 & 13.2 & 0.0 / 15.0 \\
48 & 2 & 27.0 & 10.3 & 1.7 / 18.0 \\
48 & 4 & 6.7 & 6.0 & 0.0 / 9.0 \\
\bottomrule
\end{tabular}
\caption{Potential count activations in generations without detected loops (\% with at least one activation). Structured $N=285/302$, prose $N=180/167$ for Qwen/Llama; denominators are fixed across the grid.}
\label{tab:fpgrid}
\end{table}

\paragraph{Near-loops.}
To measure approximate repetition, we decoded all 8,280 structured generations and computed two text-level measures: the zlib compression ratio (lower means more repetitive) and the share of lines that are at least 90\% similar (\texttt{difflib}) to an earlier line, ignoring lines under five characters.
Going from DRY to DRY-C, the share of generations with no verbatim loop but more than half of their lines near-duplicated rises a little, from 8.9\% to 12.2\% on Qwen and from 17.8\% to 19.7\% on Llama.
Overall repetition does not rise: the median compression ratio goes up (0.339 to 0.347 on Qwen, 0.323 to 0.347 on Llama), and the median near-duplicate line share is essentially unchanged (0.083 to 0.081; 0.050 to 0.054).
Counting a generation as repetitive when it has a verbatim loop or more than half of its lines near-duplicated, the share falls from 21.1\% (DRY) to 14.4\% (DRY-C) on Qwen and from 26.9\% to 21.4\% on Llama.
A few verbatim loops turn into drifting near-repeats, and the total share of repetitive generations falls.

\section{Structured Tasks with Reference Answers}\label{app:struct}

To score content, we built 24 tasks whose answer is fully determined by rules stated in the prompt (lookups, arithmetic, joins, sorting, rotations, a state machine): 16 ask for JSON, 4 for CSV and 4 for a Markdown table, 12 in English and 12 in Chinese, with 24 to 64 records each.
The task set, scoring and analysis were fixed before any generation.
A generation scores the F1 of complete records, $2T/(N+P)$, where $N$ is the number of reference records, $P$ the number of produced records and $T$ the number matching a reference record in every field; output that does not parse, an unclosed reasoning block or a missing answer scores 0.
Each model ran 15 samples per task and setting with the default sampler, DRY and DRY-C, and a 3{,}000-token limit; per-task means are averaged over tasks.

\begin{table}[h]
\centering\footnotesize
\setlength{\tabcolsep}{2.5pt}
\begin{tabular}{@{}llrrrrr@{}}
\toprule
Model & Sampler & F1 & Valid & Exact & No ans. & Loops \\
\midrule
Qwen3-4B & default & 31.3 & 38.9 & 25.6 & 23.6 & 1 \\
 & + DRY & 18.6 & 24.7 & 12.5 & 38.3 & 2 \\
 & + DRY-C & 20.9 & 25.6 & 13.6 & 39.2 & 0 \\
Llama-3.1-8B & default & 31.5 & 56.7 & 11.7 & 0.0 & 0 \\
 & + DRY & 24.8 & 52.2 & 5.8 & 0.0 & 0 \\
 & + DRY-C & 22.9 & 50.3 & 6.4 & 0.0 & 0 \\
\bottomrule
\end{tabular}
\caption{Structured tasks with reference answers (\%, 360 generations per row; loops are counts). Valid: the output parses with the requested header or schema; exact: every record matches in the requested order; no ans.: the reasoning block is not closed within the limit or no answer follows it.}
\label{tab:struct}
\end{table}

DRY-C minus DRY in F1 is $+2.3$ points on Qwen3-4B (95\% CI $[-0.9,5.5]$, $t$ interval stratified by format) and $-1.9$ on Llama-3.1-8B ($[-5.8,2.0]$); a stratified prompt bootstrap gives $[-0.3,5.0]$ and $[-5.5,1.3]$.
Loops are rare on these tasks, so F1 here measures correctness.
On Qwen3-4B, 24\% to 39\% of generations are still reasoning when the limit is reached.

Six of the tasks, and four further CSV and Markdown tasks run separately, require a fixed note of more than 25 tokens to be copied into every record.
With the default sampler these ten tasks reach a mean F1 of 37.8\% (Qwen) and 24.8\% (Llama); under DRY and under DRY-C every one of them scores 0 on both models.
The note contains no breaker, so DRY's length term already prevents verbatim copying, and the models insert extra spaces into later copies to avoid the match.
The count term also penalizes this copying (Appendix~\ref{app:reason}); its cost on repeated blocks dense in breakers is in Table~\ref{tab:strict}.

\section{Reasoning Benchmarks and the Count Term Alone}\label{app:reason}

\paragraph{Reasoning accuracy.}
We ran the 1,319 test problems of GSM8K \citep{cobbe2021gsm8k} and MATH-500 \citep{hendrycks2021math,lightman2023verify}, each with the prompt suffix ``Please reason step by step, and put your final answer within \texttt{\textbackslash boxed\{\}}.'' Each setting generates once per problem with the same seed across settings and up to 8,192 new tokens.
The answer is the last box; on Qwen3-4B the reasoning block must close first. MATH-500 answers are compared with \texttt{math-verify}. Intervals are paired bootstraps over problems (20,000 resamples).

\begin{table*}[t]
\centering\scriptsize
\setlength{\tabcolsep}{4pt}
\begin{tabular}{@{}llrrrrrrr@{}}
\toprule
Model & Benchmark & Default & DRY & DRY-C & Count only & DRY $-$ default & DRY-C $-$ DRY & Count only $-$ default \\
\midrule
Qwen3-4B & GSM8K & 94.4 & 91.9 & 92.3 & 93.8 & $-2.5$ $[-3.6,-1.4]$ & $+0.4$ $[-0.8,1.5]$ & $-0.6$ $[-1.5,0.2]$ \\
 & MATH-500 & 83.0 & 68.8 & 69.4 & 83.2 & $-14.2$ $[-18.0,-10.4]$ & $+0.6$ $[-2.6,3.8]$ & $+0.2$ $[-2.2,2.6]$ \\
Llama-3.1-8B & GSM8K & 83.4 & 78.3 & 78.3 & 83.2 & $-5.1$ $[-7.4,-2.8]$ & $0.0$ $[-1.9,1.9]$ & $-0.2$ $[-2.1,1.8]$ \\
 & MATH-500 & 50.2 & 38.8 & 39.4 & 49.2 & $-11.4$ $[-16.0,-6.8]$ & $+0.6$ $[-3.6,4.6]$ & $-1.0$ $[-4.8,2.8]$ \\
\bottomrule
\end{tabular}
\caption{Accuracy (\%) on reasoning benchmarks with 95\% paired bootstrap intervals in points. Count only: DRY-C without DRY's length term.}
\label{tab:reason}
\end{table*}

In our experiments DRY-C has no reasoning-accuracy loss relative to DRY.
DRY itself loses 2.5 to 14.2 points against the default sampler (Table~\ref{tab:reason}).
Its length term penalizes a match from three tokens on, and derivations reuse expressions constantly.
On Qwen MATH-500, 58 of the 85 problems that DRY loses hit the token limit; on Qwen GSM8K, 14 of 45. Others end in a miscopied answer: on one MATH-500 problem Qwen derives $(a+5)(b+2)$ and then boxes \texttt{(a + 5)(a + 2)}. On Llama most lost problems end in a wrong boxed answer (83 of 157 on GSM8K, 82 of 102 on MATH-500), and the rest without a boxed answer.
Scoring unboxed answers from the end of the text keeps the pattern: DRY minus default becomes $-2.6$, $-13.8$, $-3.6$ and $-10.8$ in table order, and DRY-C minus DRY $+0.4$, $+1.0$, $-0.8$ and $+0.4$.

\citet{weidmann2026dry} report GSM8K accuracy of 92.1\% without DRY and 92.0\% with it on Llama-3-70B, with a 1,024-token range.
On these smaller models and on harder problems the cost is large.
Setting our range to 1,024 changes MATH-500 by $+0.8$ points on Llama ($[-3.2,4.8]$) and $+5.0$ on Qwen ($[1.8,8.2]$), which leaves Qwen 9.2 points below default ($[-12.6,-5.8]$).

\paragraph{The count term alone.}
With the length term removed, accuracy returns to the default level on all four benchmark rows, and no benchmark generation loops to the limit.
Table~\ref{tab:count} gives the structured prompts: 92 of 1,800 generations contain a loop (DRY-C: 34; DRY: 167) and one reaches the limit, a 19-token unit on Phi-4-mini that starts 99 tokens before it.
The length term also pushes generations toward new items: on four of the five models, coverage with the count term alone is 4.2 to 6.5 points below DRY, and duplication is higher on all five. Coverage still exceeds the default sampler's on all five.

\begin{table}[t]
\centering\scriptsize
\setlength{\tabcolsep}{2.2pt}
\begin{tabular}{@{}lrrrrrrrr@{}}
\toprule
 & \multicolumn{2}{c}{Limit} & \multicolumn{2}{c}{Cov.} & \multicolumn{2}{c}{Dup.} & \multicolumn{2}{c}{JSON} \\
Model & DRY & Count & DRY & Count & DRY & Count & DRY & Count \\
\midrule
Qwen3-4B & 31 & 0 & 40.5 & 36.3 & 18.5 & 21.9 & 31.1 & 44.4 \\
Llama-3.1-8B & 32 & 0 & 71.5 & 66.6 & 12.9 & 19.4 & 26.7 & 26.7 \\
Qwen3-8B & 34 & 0 & 47.2 & 42.2 & 13.7 & 17.3 & 40.0 & 42.2 \\
Gemma-2-9B & 0 & 0 & 45.6 & 46.7 & 2.1 & 3.5 & 55.6 & 44.4 \\
Phi-4-mini & 53 & 1 & 44.5 & 38.0 & 26.3 & 33.4 & 33.3 & 51.1 \\
\bottomrule
\end{tabular}
\caption{DRY against the count term alone on structured prompts, 360 generations per cell. Limit: loops reaching the length limit; other columns as in Table~\ref{tab:quality}.}
\label{tab:count}
\end{table}

Requested copying recovers about half of what DRY loses: strict successes out of 160 are 76 on Qwen and 62 on Llama (default 126 and 120; DRY 29 and 17).
The four block tasks go the other way, at 2 and 9 of 40 (DRY 12 and 17).
On the 24 tasks with reference answers, F1 is 26.5\% on Qwen and 26.8\% on Llama. That is $+7.9$ ($[0.7,15.0]$) and $+2.0$ ($[-5.0,9.0]$) against DRY, and $-4.9$ ($[-14.9,5.2]$) and $-4.7$ ($[-11.5,2.2]$) against default.
The ten fixed-note tasks score 5.5\% and 3.6\% (default 37.8\% and 24.8\%): the notes exceed $a'$, so the count term acts from the third copy.
Where DRY stays on, the count term closes its blind spot; where DRY is replaced, the count term alone keeps reasoning accuracy and about half of the requested copying that DRY loses.

\end{document}